\documentclass[10pt,twocolumn,letterpaper]{article}
\usepackage{cvpr}
\usepackage{times}
\usepackage{epsfig}
\usepackage{graphicx}
\usepackage{amsmath}
\usepackage{amssymb}
\usepackage{multirow}
\usepackage{enumitem}
\usepackage{array}
\usepackage{caption} 
\usepackage{soul}
\usepackage{colortbl}
\usepackage{arydshln}

\makeatletter
\newcommand{\printfnsymbol}[1]{%
  \textsuperscript{\@fnsymbol{#1}}%
}
\usepackage[square,numbers]{natbib}
\usepackage[font=small]{caption}

\newcolumntype{P}[1]{>{\centering\arraybackslash}p{#1}}

\usepackage[pagebackref=true,breaklinks=true,letterpaper=true,colorlinks,bookmarks=false]{hyperref}
\usepackage[capitalise]{cleveref}                 
\crefname{equation}{}{}                           
\Crefname{equation}{}{}                           

\cvprfinalcopy 

\def\cvprPaperID{6771} 

\ifcvprfinal\fi
\begin{document}

\title{SCATTER: Selective Context Attentional Scene Text Recognizer}

\author{
    Ron Litman\thanks{Authors contribute equally.} , Oron Anschel\printfnsymbol{1}, Shahar Tsiper, Roee Litman, Shai Mazor and R. Manmatha \\
    Amazon Web Services\\
    {\tt\small \{litmanr, oronans, tsiper, rlit, smazor, manmatha\}@amazon.com}
}

\newcommand{\AlgoName}{SCATTER }
\newcommand{\SelBlockName}{Selective-Contextual Refinement }
\newcommand{\SelDecoName}{Selective-Decoder }


\newcommand{\GreyHeatmap}[1]{#1}
\newcommand{\GreyHeatmapB}[1]{#1}

\maketitle
\thispagestyle{empty}

\begin{abstract}
Scene Text Recognition (STR), the task of recognizing text against complex image backgrounds, is an active area of research.
Current state-of-the-art (SOTA) methods still struggle to recognize text written in arbitrary shapes.
In this paper, we introduce a novel architecture for STR, named Selective Context ATtentional Text Recognizer (SCATTER).
SCATTER utilizes a stacked block architecture with intermediate supervision during training, that paves the way to successfully train a deep BiLSTM encoder, thus improving the encoding of contextual dependencies.
Decoding is done using a two-step 1D attention mechanism.
The first attention step re-weights visual features from a CNN backbone together with contextual features computed by a BiLSTM layer.
The second attention step, similar to previous papers, treats the features as a sequence and attends to the intra-sequence relationships.
Experiments show that the proposed approach surpasses SOTA performance on irregular text recognition benchmarks by 3.7\% on average.
\end{abstract}

\section{Introduction}
\label{sec:introduction}

We address the task of reading text in natural scenes, commonly referred to as Scene Text Recognition (STR).
Although STR has been active since the late 90's, only recently accuracy reached a level that enables commercial applications, this is mostly due to advances in deep neural networks research for computer vision tasks.
Applications for STR include, among others, recognizing street signs in autonomous driving, company logos, assistive technology for the blind and translation apps in mixed reality.

\begin{figure}[ht!]
  \centering
  \includegraphics[width=\columnwidth]{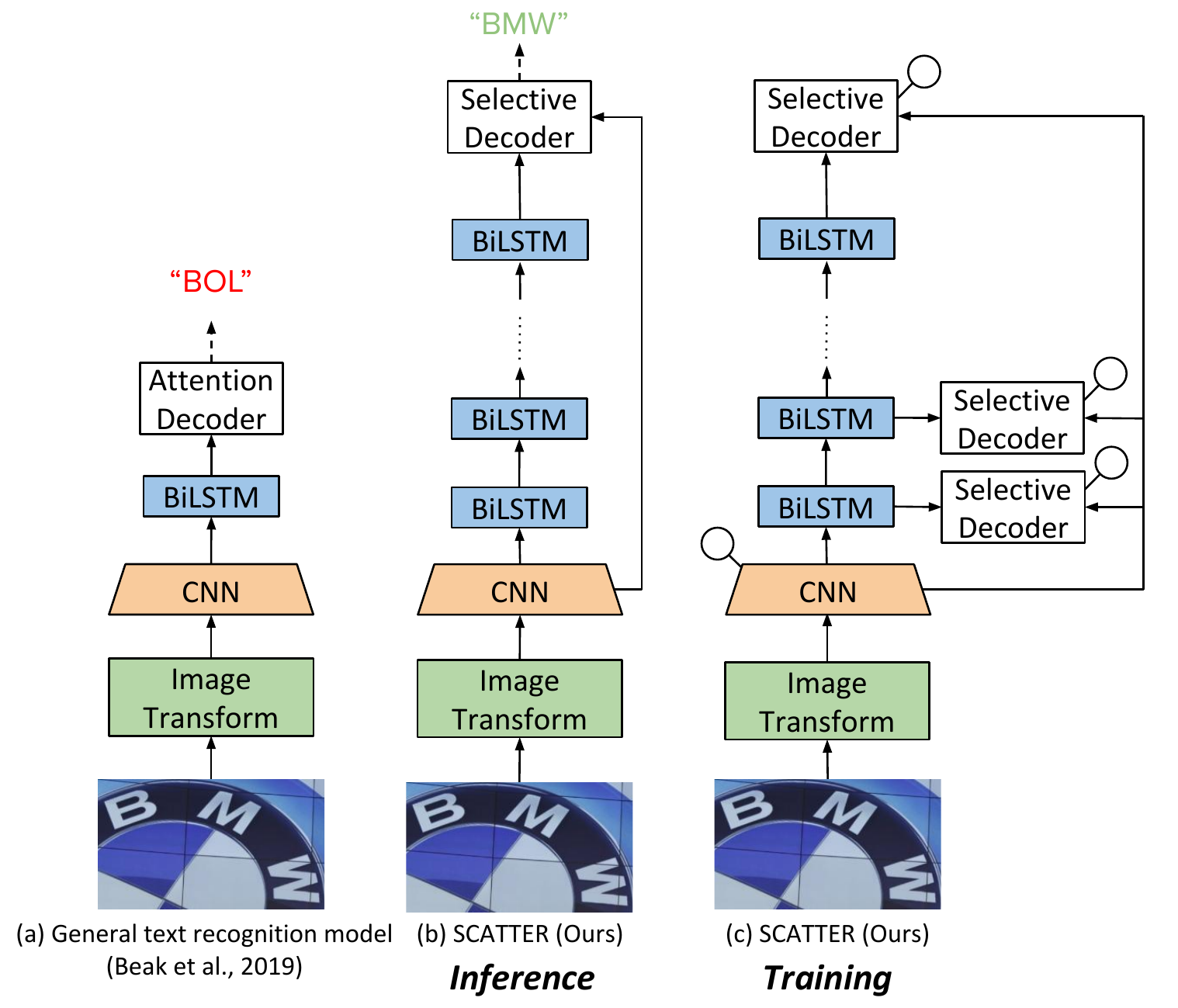}
  \caption{\textbf{The proposed \AlgoName training and inference architecture}.
  We introduce intermediate supervision combined with selective-decoding to stabilize the training of a deep BiLSTM encoder (The circle represents where a loss function is applied).
  Decoding is done using a selective-decoder that operates on visual features from the CNN backbone and contextual features from the BiLSTM encoder, while employing a two-step attention.}
  \label{fig:high_level_method}
\end{figure}

Text in natural scenes is characterised by a large variety of backgrounds, and arbitrary imaging conditions that can lead to low contrast, blur, distortion, low resolution, uneven illumination and other phenomena and artifacts.
In addition, the sheer magnitude of possible font types and sizes add another layer of difficulty that STR algorithms must overcome. 
Generally, recognizing scene text can be divided into two main tasks - text detection and text recognition.
Text detection is the task of identifying the regions in a natural image, that contain arbitrary shapes of text.
Text recognition deals with the task of decoding a cropped image that contains one or more words into a digital string of its contents.

In this paper, we propose a method for text recognition; we assume the input is a cropped image of text taken from a natural image, and the output is the recognized text string within the cropped image.
As categorized by previous works \cite{Baek2019clova, Wang2019sar}, text images can be divided into two categories:
\textit{Irregular text} for arbitrarily shaped text (e.g. curved text), as seen in \cref{fig:high_level_method}, and \textit{regular text} for text with nearly horizontally aligned characters (examples are provided in the supplementary material).

Traditional text recognition methods~\cite{Wang2011bottom, Wang2010bottom, shi2014bottom} detect and recognize text character by character, however, these methods have an inherent limitation -- they do not utilize sequential modeling and contextual dependencies between characters.

Modern methods treat STR as a sequence prediction problem.
This technique alleviates the need for character-level annotations (per-character bounding box) while achieving superior accuracy.
The majority of these sequence-based methods rely on Connectionist Temporal Classification (CTC) \cite{Bai2015crnn, Gao2019dsan}, or attention-based mechanisms \cite{Bai2018aster, Wang2019sar}.
Recently, Baek et al. \cite{Baek2019clova} proposed a modular four-step STR framework, where the individual components are interchangeable allowing for different algorithms.
This modular framework, along with its best performing component configuration, is depicted in \cref{fig:high_level_method} (a).
In this work, we build upon this framework and extend it.

While accurately recognizing regular scene text remains an open problem, recent irregular STR benchmarks (e.g., ICD15, SVTP) have shifted research focus to the problem of recognizing text in arbitrary shapes.
For instance, Sheng et al.~\cite{Sheng209nrtr} adopted the Transformer~\cite{Vaswani2017trans} model for STR, leveraging the transformers ability to capture long-range contextual dependencies.
The authors in \cite{Wang2019sar} passed the visual features from the CNN backbone through a 2D attention module down to their decoder.
Mask TextSpotter~\cite{liao2019mask} unified the detection and the recognition tasks with a shared backbone architecture.
For the recognition stage, two types of prediction branches are used, and the final prediction is selected based on the output of the more confident branch.
The first branch uses semantic segmentation of characters, and requires additional character-level annotations.
The second branch employs a 2D spatial attention-decoder.

Most of the aforementioned STR methods perform a sequential modeling step using a recursive neural network (RNN) or other sequential modeling layers (e.g., multi-head attention \cite{Sheng209nrtr}), usually in the encoder and/or the decoder.
This step is performed to convert the \textbf{visual feature} map into a \textbf{contextual feature} map, which better captures long-term dependencies.
In this work, we propose using a stacked block architecture for repeated feature processing, a concept similar to that used in other computer-vision tasks such as in~\cite{wei2016cpm} and later in~\cite{Newell2016hour, Newell2017hour}.
The authors above showed that repeated processing used in conjunction with intermediate supervision could be used to increasingly refine predictions. 

\begin{figure}[t]
  \centering
    \includegraphics[width=0.66\columnwidth]{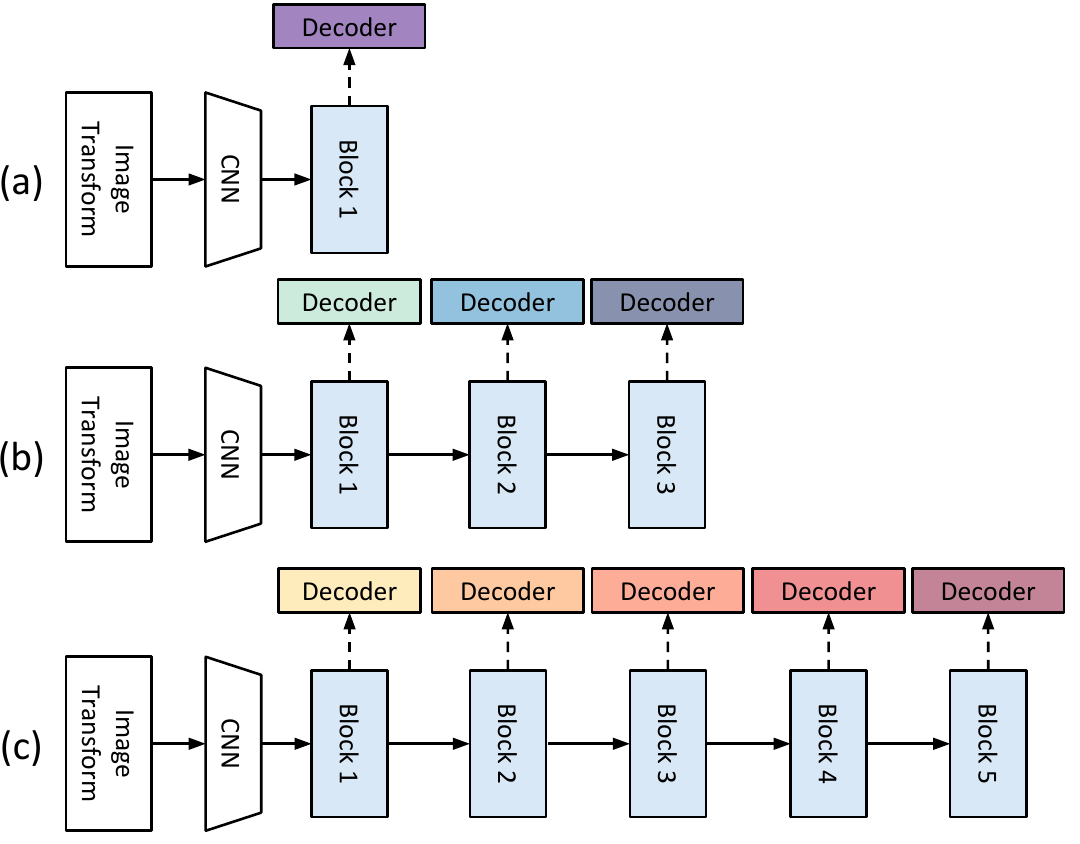}
    \includegraphics[width=0.33\columnwidth]{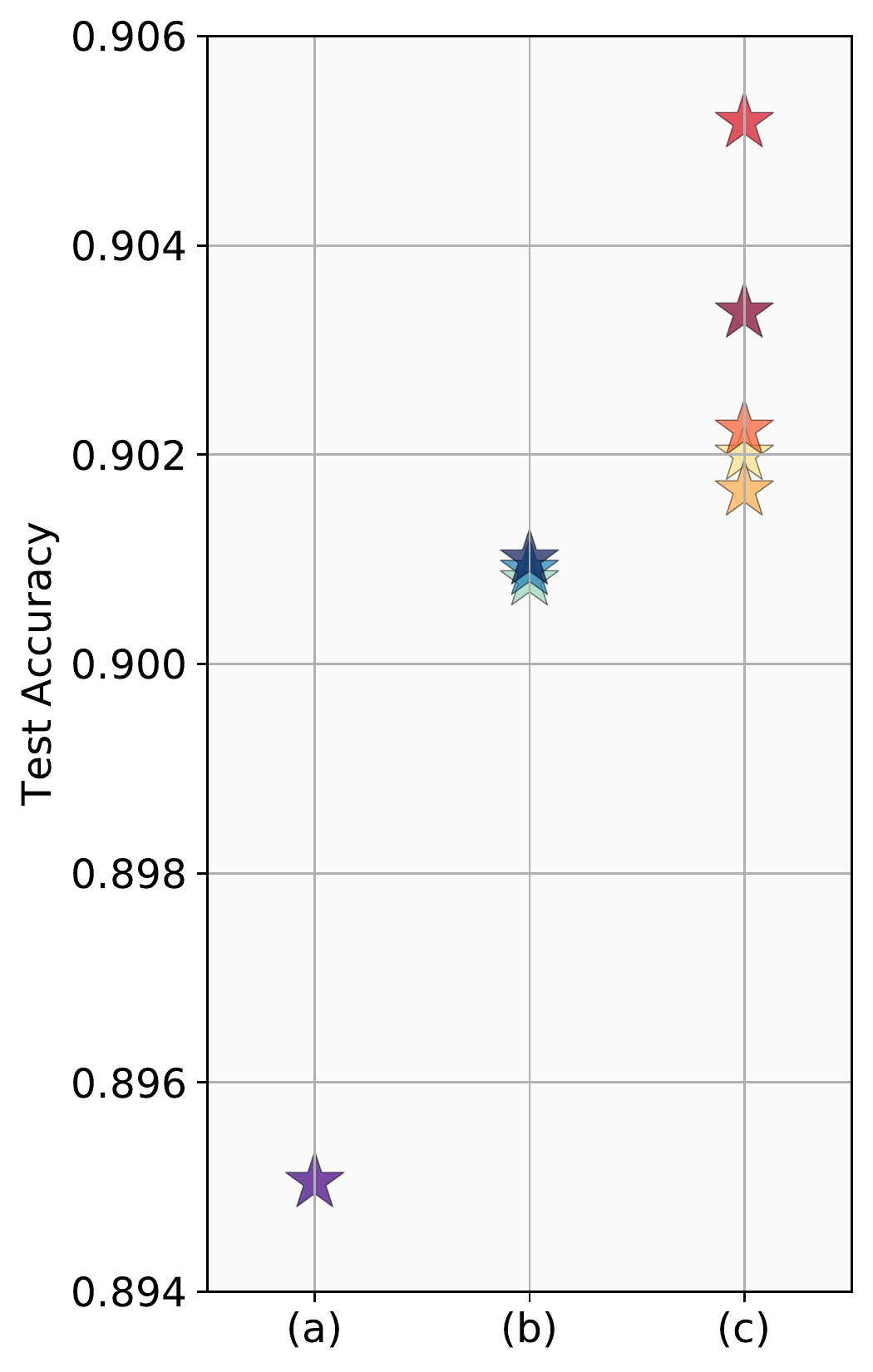}
  \caption{Average test accuracy (IIT5K, SVT, IC03, IC13,IC15, SVTP, CUTE) at intermediate decoding steps, compared across different network depths used in training.
  Given a computation budget, improved results can be obtained by first training a deeper network, and then running only the first decoder(s) during inference.}
  \label{fig:val_acc_heads}
\end{figure}

In this paper, we propose the \textbf{S}elective \textbf{C}ontext \textbf{AT}tentional \textbf{TE}xt \textbf{R}ecognizer (SCATTER) architecture.
Our method, as depicted in \cref{fig:high_level_method}, utilize a stacked block architecture for repetitive processing with intermediate supervision in training, and a novel selective-decoder.
The selective-decoder receives features from two different layers of the network, namely, visual features from a CNN backbone and contextual features computed by a BiLSTM layer, while using a two-step 1D attention mechanism.
\Cref{fig:val_acc_heads} shows the accuracy levels computed at the intermediate auxiliary decoders, for different stacking arrangements, thus demonstrating the increase in performance as additional blocks are added in succession.
Interestingly, training with additional blocks in sequence leads to an improvement in the accuracy of the intermediate decoders as well (compared to training with a shallower stacking arrangement).

This paper presents two main contributions:
\begin{enumerate}[nolistsep]
\item We propose a repetitive processing architecture for text recognition, trained with intermediate selective decoders as supervision.
Using this architecture we train a deep BiLSTM encoder, leading to SOTA results on irregular text.
\item A selective attention decoder, that simultaneously decodes both visual and contextual features by employing a two-step attention mechanism.
The first attention step figures out which visual and contextual features to attend to.
The second step treats the features as a sequence and attends the intra-sequence relations.
\end{enumerate}

\section{Related Work}
\label{sec:related_work}

STR has attracted considerable attention over the past few years~\cite{Zisserman2015large, Hu2017grccn, fedor2018rosetta, Liu20168stn}.
Comprehensive surveys for scene text detection and recognition may be found in~\cite{Ye2014survey, Baek2019clova, Long2018survey}.
As mentioned above, STR may be divided into two categories: regular and irregular texts (further examples are provided in the supplementary material).
Earlier papers~\cite{Wang2011bottom, Wang2010bottom, shi2014bottom}, focused on regular text and used a bottom-up approach, which involved segmenting individual characters with a sliding window, and then recognizing the characters using hand-crafted features.
A notable issue with the bottom-up approaches above, is that they struggle to use contextual information; instead they rely on accurate character classifiers.
Shi et al. 2015 \cite{Bai2015crnn} and He et al.~\cite{He2016cnn} considered words as sequences of varying lengths, and employed RNNs to model the sequences without explicit character separation.
Shi et al. 2016~\cite{bai2016tps} presented a successful end-to-end trainable architecture using the sequence approach, without relying on character level annotations.
Their solution employed a BiLSTM layer to extract the sequential feature vectors from the input feature maps, these vectors are then fed into an attention-Gated Recurrent Unit (GRU) module for decoding.

The methods mentioned above introduced significant improvements in STR accuracy on public benchmarks.
Therefore, recent work has shifted focus to the more challenging problem of recognizing irregularly shaped text, hence promoting new lines of research. 
Topics such as input rectification, character-level segmentation, 2D attentional feature maps and self attention have emerged, pushing the envelope on irregular STR.
Shi et al. 2018~\cite{Bai2018aster} rectified oriented or curved text based on a Spatial Transformer Network (STN).
Liu et al. 2018~\cite{Wei2018charnet} introduced a Character-Aware Neural Network (Char-Net) to detect and rectify individual characters.
A combination of a CTC-Attention mechanism within an encoder-decoder framework, that was used for speech recognition tasks, was used for STR in~\cite{Zuo2019join}, showing the benefits of joint CTC-Attention learning.
The authors in \cite{Gao2019dsan} proposed two supervision branches to tackle explicit and implicit semantic information.
In \cite{Chen2019laeg} a gate was inserted to the recurrent decoder, for controlling the transmission-weight of the previous embedded vector, demonstrating that context is not always needed for decoding.
The authors of Mask TextSpotter~\cite{liao2019mask} unified text detection and text recognition in an end-to-end fashion.
For recognition they used two separate branches, a branch that uses visual (local) features and a branch which utilizes contextual information in the form of 2D attention.

More recent approaches have proposed leveraging various attention mechanisms for improved results. 
Li et al.~\cite{Wang2019sar} combined both visual and contextual features while utilizing a 2D attention within the encoder-decoder.
Other researchers borrowed ideas from the Natural Language Processing (NLP) domain and adopted a transformer-based architecture~\cite{Vaswani2017trans}.
One of them is Sheng et al.~\cite{Sheng209nrtr}, that used a self-attention mechanism for both the encoder and the decoder.

Our method differs from above approaches, by being the first to utilize a stacked block architecture for text recognition.
Namely, we show that repetitive processing for text recognition, trained with intermediate selective decoders as supervision (similar to ~\cite{wei2016cpm, Newell2016hour, Newell2017hour}), increasingly refines text predictions.


\section{Methodology}
\label{sec:method}

\begin{figure*}[ht!]
\normalsize
  \centering
  \includegraphics[width=\textwidth]{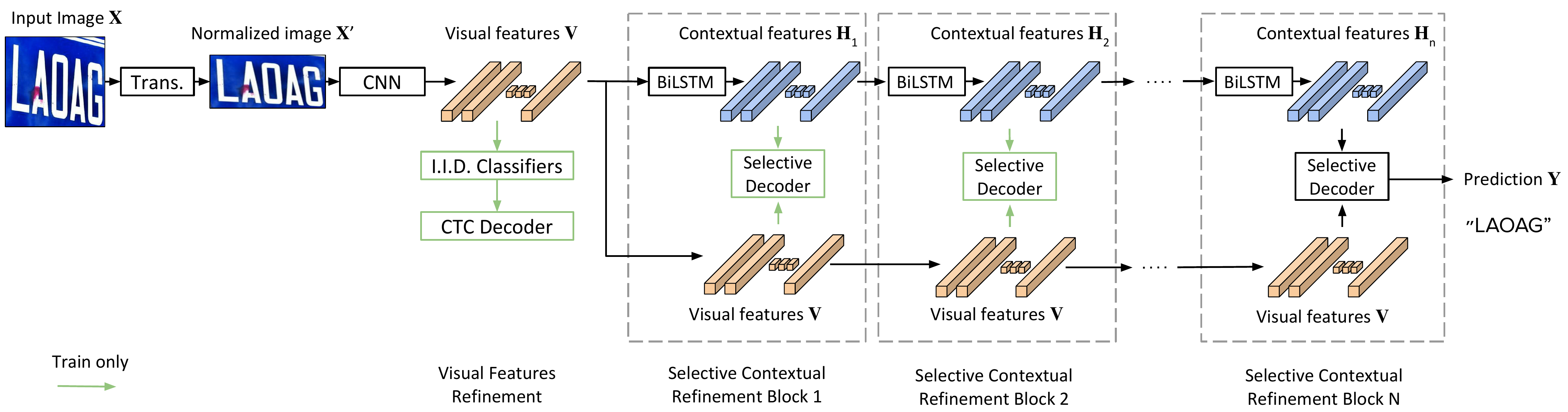}
  \caption{The proposed \AlgoName architecture introduces, context refinement, intermediate supervision (additional decoders), and a novel selective-decoder.}
  \label{fig:architecture}
\end{figure*}

As presented in \cref{fig:architecture}, our proposed architecture consists of four main components:
\begin{enumerate}[nolistsep]
\item \textbf{Transformation:} the input text image is normalized using a Spatial Transformer Network (STN)~\cite{Zisserman2018stn}.
\item \textbf{Feature Extraction:} maps the input image to a feature map representation while using a text attention module~\cite{Gao2019dsan}.
\item \textbf{Visual Feature Refinement:} provides direct supervision for each column in the visual features.
This part refines the representation in each of the feature columns, by classifying them into individual symbols.
\item \textbf{\SelBlockName Block:} Each block consists of a two-layer BiLSTM encoder that outputs contextual features.
The contextual features are concatenated to the visual features computed by the CNN backbone.
This concatenated feature map is then fed into the selective-decoder, which employs a two-step 1D attention mechanism, as illustrated in \cref{fig:selective_decoder}. 
\end{enumerate}
In this section we describe the training architecture of \AlgoName, while addressing the differences between training and inference.

\subsection{Transformation}
The transformation step operates on the cropped text image $X$, and transforms it into a normalized image $X^\prime$.
We use a Thin Plate Spline (TPS) transformation, a variant of the STN, as used in~\cite{Baek2019clova}.
TPS employs a smooth spline interpolation between a set of fiducial points.
Specifically, it detects a pre-defined number of fiducial points at the top and bottom of the text region, and normalizes the predicted region to a constant predefined size.

\subsection{Feature Extraction}
In this step a convolutional neural network (CNN) extracts features from the input image.
We use a 29-layer ResNet as the CNN's backbone, as used in~\cite{bai2017accurate}.
The output of the feature encoder is 512 channels by $N$ columns.
Specifically, the feature encoder gets an input image $X^\prime$ and outputs a feature map $F = [f_1,f_2,...,f_N]$.
Following the feature map extraction, we use a text attention module, similar to~\cite{Gao2019dsan}.
The attentional feature map can be regarded as a visual feature sequence of length $N$, denoted as $V = [v_1,v_2,...,v_N]$, where each column represents a frame in the sequence.

\subsection{Visual Feature Refinement}
Here, the visual feature sequence $V$ is used for intermediate decoding.
This intermediate supervision is aimed at refining the character embedding (representations) in each of the columns of $V$, and is done using CTC based decoding.
We feed $V$ through a fully connected layer that outputs a sequence $H$ of length $N$.
The output sequence is fed into a CTC~\cite{Graves2006ctc} decoder to generate the final output.
The CTC decoder transforms the output sequence tensor into a conditional probability distribution over the label sequences, and then selects the most probable label.
The transcription procedure is given by
\begin{align}
l = B(\operatorname*{arg\,max}_\pi p(\pi|H)) \,,
\end{align}
where the probability of $\pi$ is defined as
\begin{align}
p(\pi|H) = \prod_{t=1}^{N} y^{t}_{\pi_t} \,.
\end{align}
Here $y^{t}_{\pi_t}$ is the probability of generating the character $\pi_t$ at time stamp $t$, and $B$ is a mapping function that removes all repeated characters and blanks.
The CTC algorithm assumes that the columns are conditionally independent, and at each time stamp the output is a single character probability score.
The loss for this branch, denoted by $L_\text{CTC}$, is the negative log-likelihood of the ground-truth conditional probability, as in~\cite{Bai2015crnn}.


\subsection{\SelBlockName Block}
\label{sec:contextlevel}
The features extracted by the CNN are limited to its receptive field, and may suffer due to the lack of contextual information.
To mitigate this drawback, we employ a two-layer BiLSTM~\cite{Graves2008rnn} network over the feature map $V$, outputting $H = [h_1, h_2,..., h_n]$. 
We concatenate the BiLSTM output with the visual feature map, yielding $D = (V, H)$, a new feature space. 

The feature space $D$ is used both for selective decoding, and as an input to the next \SelBlockName block.
Specifically, the concatenated output of the $j$th block can be written as $D_j = (V, H_j)$.
The next ${j+1}$ block uses $H_j$ as input to the two-layer BiLSTM, yielding $H_{j+1}$, and the $j+1$ feature space is updated such that $D_{j+1} = (V, H_{j+1})$.
The visual feature map $V$ does not undergo any further updates in the \SelBlockName blocks, however we note that the CNN backbone is updated with back-propagated gradients from all of the selective-decoders.
These blocks can be stacked together as many times as needed, according to the task or accuracy levels required, and the final prediction is provided by the decoder from the last block.

\subsubsection{Selective-Decoder}

\begin{figure}[t]
  \centering
  \includegraphics[width=\columnwidth]{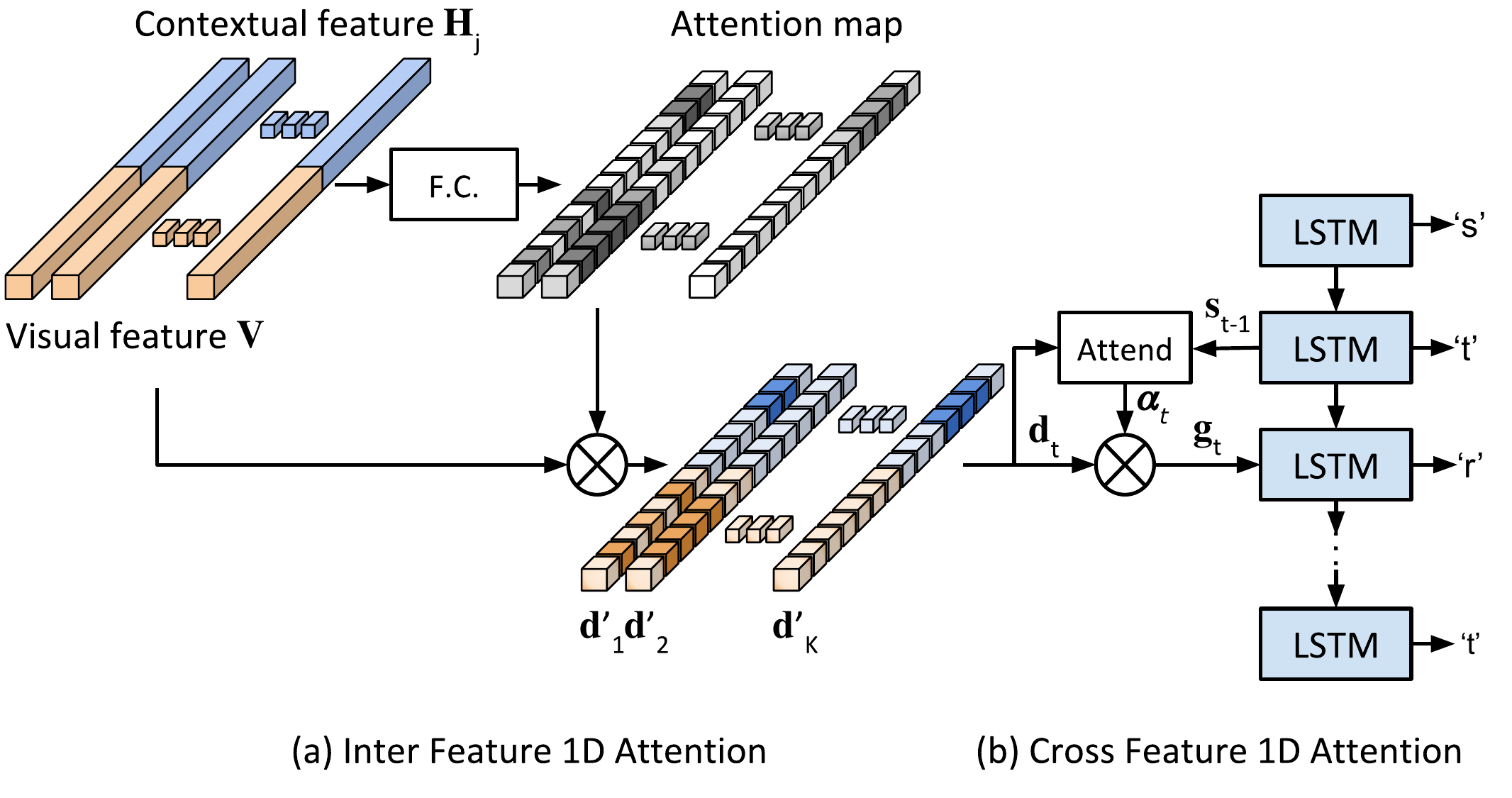}
  \caption{Architecture of the Two-Step Attention Selective-Decoder.}
  \label{fig:selective_decoder}
\end{figure}

We employ a two-step attention mechanism, as illustrated in \cref{fig:selective_decoder}.
First, we use a 1D self attention operating on the features $D$.
A fully connected layer is used to compute an attention map from these features.
Next, an element-wise product is computed between the attention map and $D$, yielding the attentional features $D^\prime$.
The decoding of $D^\prime$ is done with a separate attention-decoder, such that for each $t$-time-step the decoder outputs $y_t$, similar to \cite{bai2017accurate, bai2018edit}.

Decoding starts by computing the vector of attentional weights, $\alpha_t \in R^N$:
\begin{align}
\label{equ:equ2}
e_{t,i} & = w^T \tanh(Ws_{t-1} + Vd^\prime_i + b) \\
\alpha_{t,i} & = \exp (e_{t,i}) / \sum_{i^*=1}^{n}e_{t,i^*} \,,
\end{align}
where $b, w, W, V$ are trainable parameters, $s_{t-1}$ is the hidden state of the recurrent cell within the decoder at time $t$, and $d\prime$ is a column of $D\prime$.
The decoder linearly combines the columns of $D^\prime$ into a vector $G$, which is called a glimpse:
\begin{align}
g_{t} = \sum_{i=1}^{n}\alpha_{t,i}d^\prime_i \,.
\label{equ:equ4}
\end{align}
Next, a recurrent cell of the decoder is fed with
\begin{align}
(x_t,s_t) = \text{RNN}\Big(s_{t-1}, \big(g_t,f(y_{t-1})\big)\Big) \,,
\end{align}
where $(g_t,f(y_{t-1}))$ denotes the concatenation between $g_t$ and the one-hot embedding of $y_{t-1}$.

The probability for a given character $p(y_t)$ can now be recovered by:
\begin{align}
p(y_{t}) = \text{softmax} (W_o x_t + b_o) \,.
\end{align}
The loss for the $j$th block is the negative log-likelihood, denoted as $L_{\text{Attn}, j}$, as in~\cite{bai2017accurate, bai2018edit}.


\subsection{Training Losses}
The objective function is given by
\begin{align}
L = \lambda_{\text{CTC}} \cdot L_\text{CTC} + \sum_{j=1}\lambda_j L_{\text{Attn}, j} \,,
\end{align}
where $L_\text{CTC}$ is the loss function of the CTC decoder and $\sum_{j=1}\lambda_j L_{\text{Attn}, j}$ is the sum of the losses from all of the \SelBlockName blocks, as defined above.
The $\lambda$ notation depicts a hyper-parameter used to balance the trade-off between the different supervisions, and specifically, $\lambda_{CTC}, \lambda_j$ are empirically set to 0.1, 1.0 respectively for all $j$.

\subsection{Inference}
Once training is done, for test time we remove all of the intermediate decoders, as they are used only for additional supervision and refinement of intermediate feature.
The visual features $V$ are processed by the BiLSTM layers in all blocks, and are also fed directly, via a skip connection, to the final selective-decoder. 
The final selective decoder is used to predict the output sequence of characters.
A visualization of these changes can be seen in \cref{fig:architecture}, where all the green colored operations are disabled during inference, and in \cref{fig:high_level_method} (b).

\section{Experiments}
\label{sec:experiments}

In this section we empirically demonstrate the effectiveness of our proposed framework.
We begin with a brief discussion regarding the datasets used for training and testing, and then describe our implementation and evaluation setup.
Next, we compare our model against state-of-the-art methods on public benchmark datasets, including both regular and irregular text.
Finally, we address the computational cost of our method.

\newcommand\dunderline[3][-1pt]{{
  \setbox0=\hbox{#3}
  \ooalign{\copy0\cr\rule[\dimexpr#1-#2\relax]{\wd0}{#2}}}}
\begin{table*}
\small
\begin{center}
\bgroup
\def\arraystretch{1.1}
\centering
  \begin{center}
  \footnotesize
  \captionsetup[table]{skip=5pt}
    \caption{Scene text recognition accuracies (\%) over seven public benchmark test datasets (number of words in each dataset are shown below the title). No lexicon is used.  In each column, the best performing result is shown in \textbf{bold} font, and the second best result is shown with an \underline{underline}. Average columns are weighted (by size) average results on the regular and irregular datasets. "*" indicates using both word-level and character-level annotations for training.}
    \label{tab:sota_results}
    \footnotesize
    \begin{tabular}{|P{4.0cm}|P{1cm}P{0.75cm}P{0.75cm}P{0.75cm}|P{1.0cm}|P{0.75cm}P{1cm}P{1cm}|P{1.0cm}|} 
      \hline
      \multirow{3}{*}{\textbf{Method}} & \multicolumn{5}{c|}{\textbf{Regular test dataset}} & \multicolumn{4}{c|}{\textbf{Irregular test dataset}}\\
      & IIIT5K & SVT & IC03 & \multicolumn{1}{c}{IC13} & \textbf{Average} & IC15 & SVTP & \multicolumn{1}{c}{CUTE} & \textbf{Average}\\
      & 3000 & 647 & 867 & \multicolumn{1}{c}{1015} & 5529 & 2077 & 645 & \multicolumn{1}{c}{288} & 3010\\
      \hline
      CRNN (2015) \cite{Bai2015crnn} & 78.2 & 80.8 & 89.4 & 86.7 & 81.8 & - & - & -& - \\
      \hline
      FAN (2017) \cite{bai2017accurate}* & 87.4 & 85.9 & 94.2 & 93.3 & 89.4 & 70.6 & - & - & -\\
      \hline
      Char-Net (2018) \cite{Wei2018charnet}* & 83.6 & 84.4 & 91.5 & 90.8 & 86.2 & 60.0 & 73.5 & - & -\\
      \hline
      AON (2018) \cite{Zhanzhan2018aon} & 87.0 & 82.8 & 91.5 & - & - & 68.2 & 73.0 & 76.8 & 70.0\\
      \hline
      EP (2018) \cite{bai2018edit}* & 88.3 & 87.5 & 94.6 & 94.4 & 90.3 & 73.9 & - & - & -\\
      \hline
	  NRTR (2018) \cite{Bai2018aster} & 86.5 & 88.3 & 95.4 & \dunderline{0.7pt}{94.7} & 89.6 & - & - & - & -\\
      \hline
      Liao et al. (2019) \cite{liao2019scene} & 91.9 & 86.4 & - & 86.4 & - & - & - & 79.9 & -\\
      \hline
      Baek et al. (2019) \cite{Baek2019clova} & 87.9 & 87.5 & 94.9 & 92.3 & 89.8 & 71.8 & 79.2 & 74.0 & 73.6\\
      \hline
      ASTER (2019) \cite{Bai2018aster} & 93.4 & 89.5 & 94.5& 91.8 & 92.8 & 76.1 & 78.5 & 79.5 & 76.9\\
      \hline
      SAR (2019) \cite{Wang2019sar} & 91.5 & 84.5 & - & 91.0 & - & 69.2 & 76.4 & 83.3 & 72.1\\
      \hline
      ESIR (2019) \cite{Zhan2019esir} & 93.3 & 90.2 & - & 91.3 & - & 76.9 & 79.6 & 83.3 & 78.1\\
      \hline
      MORAN (2019) \cite{Luo2019moran} & 91.2 & 88.3 & 95.0 & 92.4 & 91.7 & 68.8 & 76.1 & 77.4 & 71.2\\
      \hline
      Yang et al. (2019) \cite{yang2019symmetry} & \dunderline{0.7pt}{94.4} & 88.9 & 95.0 & 93.9 & 93.7 & 78.7 & 80.8 & \dunderline{0.7pt}{87.5} & 79.9\\
      \hline
      Mask TextSpotter (2019) \cite{liao2019mask}* & \textbf{95.3} & \dunderline{0.7pt}{91.8} & 95.0 & \textbf{95.3} & \textbf{94.8} & 78.2 & 83.6 & \textbf{88.5} & 80.0\\
      \hline
      \hline
      \AlgoName (1 Block) & 92.9 & 89.2 & \dunderline{0.7pt}{96.5} & 93.8 & 93.2 & 81.8 & 84.5 & 85.1 & 82.7 \\
      \hline
      \AlgoName (2 Block)& 93.5 & 89.2 & 95.9 & \dunderline{0.7pt}{94.7} & 93.6 & 81.5 & 86.2 & 86.8 & 83.0\\
      \hline
      \AlgoName (3 Block) & 93.9 & 89.3 & 96.1 & 94.6 & 93.7 & \textbf{82.8} & 85.7 & 83.7 & 83.4\\
      \hline
      \AlgoName (4 Block) & 93.4 & 90.3 & \textbf{96.6} & 94.3 & 93.7 &  82.0 & \textbf{87.0} & 86.5 & \dunderline{0.7pt}{83.5}\\
      \hline
      \AlgoName (5 Block) & 93.7 & \textbf{92.7} & 96.3 & 93.9 & \dunderline{0.7pt}{94.0} &  \dunderline{0.7pt}{82.2} & \dunderline{0.7pt}{86.9} & \dunderline{0.7pt}{87.5} & \textbf{83.7}\\
      \hline
    \end{tabular}
  \end{center}
\egroup
\end{center}
\end{table*}

\subsection{Datasets}
In this work, all SCATTER models are trained on three synthetic datasets.
The model is evaluated on four regular scene-text datasets: ICDAR2003, ICDAR2013, IIIT5K, SVT, and three irregular text datasets: ICDAR2015, SVTP and CUTE.

The training dataset is a union of three datasets:

\textbf{MJSynth} (MJ) \cite{Zisserman2014mj} is a synthetic text in image dataset which contains 9 million word box images, generated from a lexicon of 90K English words.

\textbf{SynthText} (ST) \cite{Zisserman206st} is a synthetic text in image dataset, designed for scene-text detection, and recognition.
We use a variant of the SynthText dataset composed of 5.5M samples, as used in \cite{Baek2019clova}.
This variant does not include any non-alphanumeric characters.

\textbf{SynthAdd} (SA) \cite{Wang2019sar} is a synthetic text in image dataset, that contains 1.2 million word box images.
This dataset was generated using the same synthetic engine as in ST, aiming to mitigate the lack of non-alphanumeric characters (e.g., punctuation marks) in the other datasets.

All experiments are evaluated on the seven real-word STR benchmark datasets described below.
As in many STR manuscripts (e.g, \cite{Bai2018aster, Baek2019clova, Wang2019sar}) the benchmark datasets are commonly divided into regular and irregular text, according to the text layout.

\textbf{Regular text} datasets include the following: \textbf{IIIT5K}~\cite{Mishra2012sj} consists of 2000 training and 3000 testing images that are cropped from Google image searches.
\textbf{SVT}~\cite{Wang2011bottom} is a dataset collected from Google Street View images and contains 257 training and 647 testing word-box cropped images.
\textbf{ICDAR2003}~\cite{Lucas2003ic03} contains 867 word-box cropped images for testing.
\textbf{ICDAR2013}~\cite{Karatzas2013ic13} contains 848 training and 1,015 testing word-box cropped images.

\textbf{Irregular text} datasets include the following: \textbf{ICDAR2015}~\cite{Karatzas2015ic15} contains 4,468 training and 2,077 testing word-box cropped images, all captured by Google Glass, without careful positioning or focusing.
\textbf{SVTP}~\cite{Phan2013svtp} is a dataset collected from Google Street View images and consists of 645 cropped word-box images for testing.
\textbf{CUTE 80}~\cite{Risnumawan2014cute} contains 288 cropped word-box images for testing, many of which are curved text images.

\subsection{Implementation Details}
As baseline, we use the code of Baek et al.\footnote{https://github.com/clovaai/deep-text-recognition-benchmark}~\cite{Baek2019clova}, and our architectural changes are implemented on top of it.
All experiments are trained and tested using the PyTorch\footnote{https://pytorch.org/} framework on a Tesla V100 GPU with 16GB memory.
As for the training details, we do not perform any type of pre-training.
We train using the AdaDelta optimizer, and the following training parameters are used: a decay rate of 0.95,  gradient clipping with a magnitude of 5, a batch size of 128 (with a sampling ratio of 40\%, 40\%, 20\% between MJ, ST and SA respectively).
We use data augmentation during training, and augment 40\% of the input images, by randomly resizing them and adding extra distortion.
Each model is trained for 6 epochs on the unified training set.
For our internal validation dataset, we use the union of the IC13, IC15, IIIT, and SVT training splits, to select our best model, as done in~\cite{Baek2019clova}.
All images are resized to $32 \times 100$ during both training and testing, following common practice.
In this paper, we use 36 symbol classes: 10 digits and 26 case-insensitive letters.
As for special symbols for CTC decoding, an additional “[UNK]” and a "[blank]" token are added to the label set.
For the selective-decoders three special punctuation characters are added: “[GO]”, “[S]” and “[UNK]” which indicate the start of the sequence, the end of the sequence and unknown characters (that are not alpha-numeric), respectively.

At inference, we employ a similar mechanism to \cite{Wang2019sar, Yanga209simple, Ning209master}, where images with a height larger than their width, are rotated by 90 degrees clockwise and counter-clockwise respectively.
The rotated versions are recognized alongside the original image.
A prediction confidence score is calculated as the average decoder probabilities until the '[S]' token.
We then choose the prediction with the highest confidence score as the final prediction.
Unlike \cite{Bai2018aster, Wang2019sar, Sheng209nrtr}, we do not use beam-search for decoding, although the authors in~\cite{Wang2019sar} have reported it improves accuracy by approximately 0.5\%, due to the added latency it incurs.

\begin{table*}
\small
\begin{center}
\bgroup
\def\arraystretch{1.1}
\centering
  \begin{center}
  \footnotesize
  \captionsetup[table]{skip=10pt}
    \caption{Ablation studies by changing the model hyper-parameters.
    We refer to our re-trained model using the code of Baek et al. 2019 as Baseline.
    Using intermediate supervision helps to boost results and enables stacking more blocks. Increasing the number of blocks has positive impacts on the recognition performance. * Regular Text and Irregular Text columns are  weighted (by size) average results on the regular and irregular datasets respectively.}
    \label{tab:ablation}
    \footnotesize
    \begin{tabular}{|P{0.3cm}|P{1.2cm}|P{0.85cm}|P{1.0cm}|P{1.0cm}|P{0.7cm}|P{0.65cm}|P{0.5cm}P{0.3cm}P{0.3cm}P{0.5cm}|P{0.5cm}P{0.4cm}P{0.4cm}|P{0.85cm}|P{1.0cm}|} 
      \hline
      \multirow{2}{*}{{Sec}} & \multirow{2}{*}{{Method}} & {CTC} & {Attention} & {Selective} & {LSTM} & {N} & \multicolumn{4}{c|}{{Regular test dataset}} & \multicolumn{3}{c|}{{Irregular test dataset}} & {Regular} & {Irregular}\\
      & & {Decoder} & {Decoders} & {Decoders} & {Layers} & {Blocks} & IIIT5K & SVT & IC03 & IC13 & IC15 & SVTP & CUTE & {Text*} & {Text*}\\
      \hline
      & \cite{Baek2019clova} & 0 & 1 & - & 2 & - & 87.9 & 87.5 & 94.9 & 92.3  & 71.8 & 79.2  & 74.0 & \GreyHeatmap{89.8} & \GreyHeatmapB{73.7} \\ 
      \cline{2-16}
      & Baseline & 0  & 1 & - & 2 & - & 93.0 & 88.6 & 94.3 & 92.9  & 78.0 & 81.5  & 80.1 & \GreyHeatmap{92.7} & \GreyHeatmapB{79.1} \\
      \cline{2-16}
      \multicolumn{1}{|c|}{(a)} & Baseline & 1  & 1 & - & 2 & - & 92.7 & 90.0 & 95.0 & 93.4 & 78.3 & 83.4 & 80.0 & \GreyHeatmap{92.9}& \GreyHeatmapB{79.5} \\
      \cline{2-16}
      & \AlgoName & 0  & - & 1 & 2 & 1 & 93.1 & 89.3 & 95.7 & 93.7 & 80.6 & 84.0  & 86.5 & \GreyHeatmap{93.1} & \GreyHeatmapB{81.8} \\
      \cline{2-16}
      & \AlgoName & 1  & - & 1 & 2 & 1 & 92.9 & 89.2 & 96.5 & 93.8  & 81.8 & 84.5  & 85.1 & \GreyHeatmap{93.2} & \GreyHeatmapB{82.7}  \\
      \hline
      \hline
      & \AlgoName & 0  & - & 1 & 4 & 2 & 92.9 & 89.6 & 95.3 & 93.2 & 79.3 & 84.2  & 83.7 & \GreyHeatmap{92.9} & \GreyHeatmapB{80.1} \\
      \cline{2-16}
      \multicolumn{1}{|c|}{(b)} & \AlgoName & 0  & - & 2 & 4 & 2 &  93.1 & 90.7 & 94.9 & 93.6 & 81.6 & 84.7  & 85.8 & \GreyHeatmap{93.2} & \GreyHeatmapB{82.6} \\
      \cline{2-16}
      & \AlgoName & 1  & - & 1 & 4 & 2 & 93.0 & 90.6 & 95.5 & 92.8  & 82.0 & 84.0  & 86.8 & \GreyHeatmap{93.1} & \GreyHeatmapB{82.9} \\
      \cline{2-16}
      & \AlgoName & 1  & - & 2 & 4 & 2 & 93.5 & 89.2 & 95.9 & 94.7 &  81.5 & 86.2  & 86.8 & \GreyHeatmap{93.6} & \GreyHeatmapB{83.0} \\
      \hline
      \hline
      & \AlgoName & 1  & - & 3 & 6 & 3 & 93.9 & 89.3 & 96.1 & 94.6  & 82.8 & 85.7  & 83.7 & \GreyHeatmap{93.7} & \GreyHeatmapB{83.4} \\
      \cline{2-16}
      \multicolumn{1}{|c|}{(c)} & \AlgoName & 1  & - & 4 & 8 & 4 & 93.4 & 90.3 & 96.6 & 94.3 &  82.0 & 87.0  & 86.5 & \GreyHeatmap{93.7} & \GreyHeatmapB{83.5} \\
      \cline{2-16}
      & \AlgoName & 1  & - & 5 & 10 & 5 & 93.7 & 92.7 & 96.3 & 93.9 & 82.2 & 86.9  & 87.5 & \GreyHeatmap{94.0} & \GreyHeatmapB{83.7} \\
      \cline{2-16}
      & \AlgoName & 1  & - & 6 & 12 & 6 & 93.2 & 90.9 & 96.2 & 94.1 & 82.0 & 86.2  & 84.8 & \GreyHeatmap{93.6} & \GreyHeatmapB{83.2} \\
      \hline
    \end{tabular}
  \end{center}
\egroup
\end{center}
\end{table*}

\subsection{Comparison to State-of-the-art}

In this section, we measure the accuracy of our proposed framework on several regular and irregular scene text benchmarks while comparing the results to the latest SOTA recognition methods.
As seen in \cref{tab:sota_results}, our \AlgoName architecture with 5 blocks outperforms the current SOTA, the Mask TextSpotter~\cite{liao2019mask} algorithm, on irregular scene text benchmarks (i.e., IC15, SVTP, CUTE) by an absolute margin of 3.7\% on average.
Our approach provides an accuracy increase of \textbf{+4.0} pp (78.2\% vs. 82.2\%) on IC15, \textbf{+3.3} pp (83.6\% vs. 86.9\%) on SVTP, and is the second best to Mask TextSpotter~\cite{liao2019mask} on the CUTE (88.5\% vs. 87.5\%) dataset.
Additionally, the proposed method outperforms the other methods both on SVT and IC03 regular scene text datasets, and achieves comparable SOTA performance on the other regular scene text datasets (i.e.,IIIT5K and IC13). 

To summarize, our model achieves the highest recognition score on 4 out of 7 benchmarks, and the second best score on 2 more benchmarks.
Unlike other methods, which perform well on either regular or irregular scene text benchmarks, our approach is a top performer on all benchmarks.

We would like to briefly discuss key differences between Mask TextSpotter~\cite{liao2019mask} and this work.
The algorithm in~\cite{liao2019mask} relies on annotations that contain character level annotations, information that our algorithm does not require.
These annotations contribute an average increase of 0.9 pp, and 0.6 pp on regular and irregular text datasets as reported in the original paper.
Hence, without the character level annotations, our model achieves slightly better results on regular text (93.9\% vs. 94\%) and significantly better results on irregular text (79.4\% vs. 83.7\%).
Our approach, on the other hand, does not require these annotations, which are expensive and hard to annotate, especially for real-world data.

\begin{figure}[t]
  \centering
  \includegraphics[width=\columnwidth]{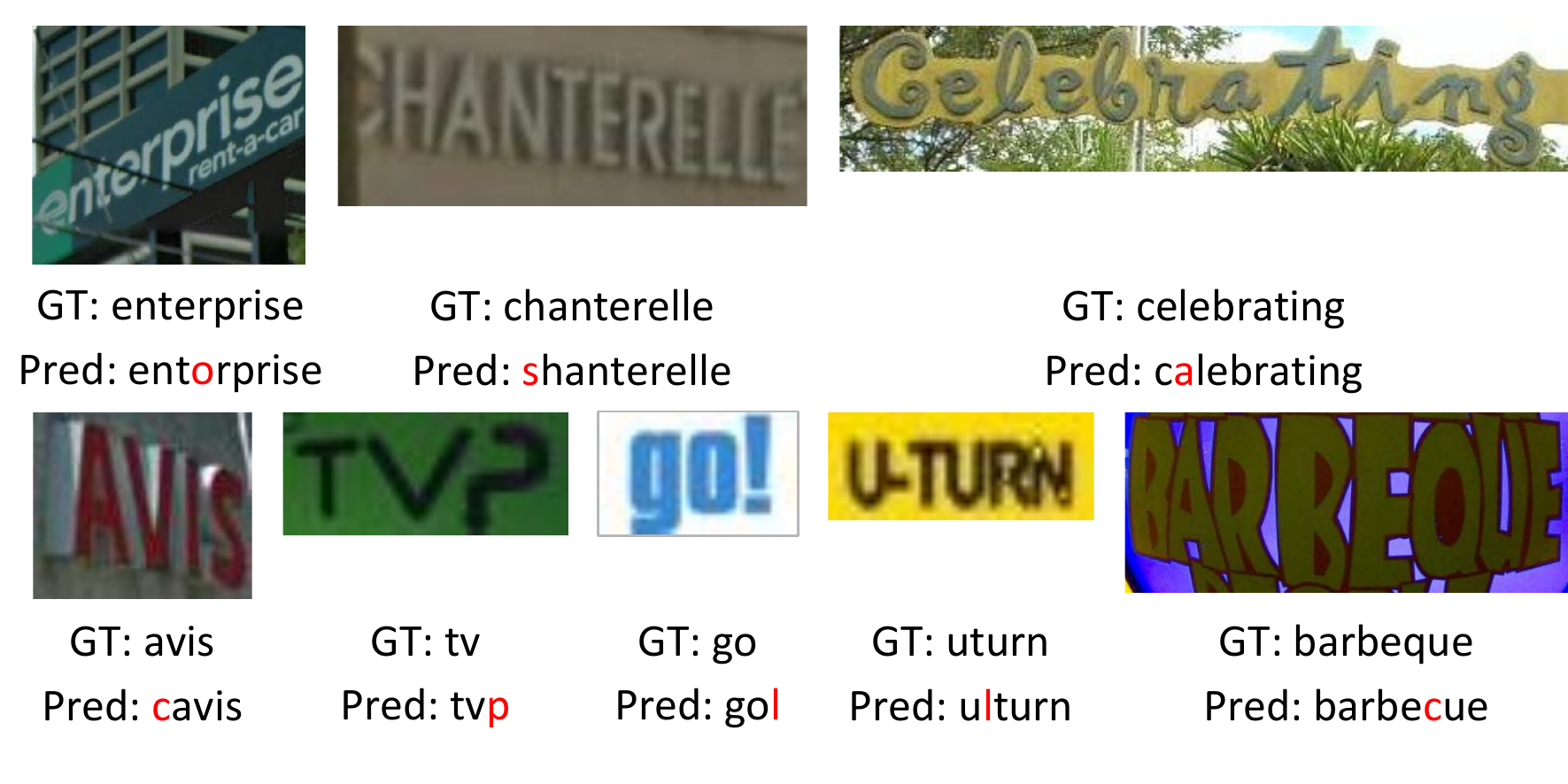}
  \caption{Failure cases of our model. “GT” stands for the ground-truth annotation, and “Pred” is the predicted result.}
  \label{fig:fail_cases}
\end{figure}

In \cref{fig:fail_cases} we display failure cases of our method.
The failure cases are mostly composed of blurry images, partial character occlusions, difficult lighting conditions, and miss-recognition of punctuation marks.

\subsection{Computational Costs}

During inference, only the selective decoder of the final block is kept active, as shown in \cref{fig:high_level_method}.
The total computational cost of our proposed architecture with a single block is 20.1 ms.
The additional computational cost of each \textit{intermediate} contextual refinement block during inference translates to a 3.2 ms per block.
For a 5 block architecture (our best setup) this translates to a total increase of 12.8 ms and a total forward pass of 32.9 ms.

Furthermore, given a computational budget on inference time, performance can be improved by training the system with a large number of blocks and pruning them for inference.
For example, an architecture trained with five blocks, and then pruned to a single block, is capable of outperforming an architecture solely trained with a single block. 
\Cref{fig:val_acc_heads}(2c) demonstrates a network trained with five blocks and the average test accuracy of intermediate decoders.
This showcases that pruning leads to an increase of \textbf{+0.4} pp and \textbf{+1.3} pp on regular an irregular datasets respectively (under the same computational budget).
This novel feature of SCATTER allows for faster inference if needed and in some cases pruning can even boost results.

\section{Ablation Experiments}
\label{sec:ablation}

In this section, we perform a series of experiments to better understand the performance improvements and analyze the impact of our key contributions.
Throughout this section, we use a weighted-average (by the number of samples) of the results on the regular and irregular test datasets.
For completeness, the first and second rows in \cref{tab:ablation} show the reported results in~\cite{Baek2019clova}, and the improved results of our re-trained model of~\cite{Baek2019clova} with our custom training settings.


\subsection{Intermediate Supervision \& Selective Decoding}
Section (a) of \cref{tab:ablation} shows an improvement in accuracy by adding the intermediate CTC supervision and the proposed selective-decoder.
Between row two and three of section (a) we add a CTC decoder for intermediate supervision which improves the baseline results by \textbf{+0.2} pp and \textbf{+0.4} pp on regular and irregular text respectively.
The fourth row demonstrates the improvement compared to the baseline results by replacing the standard attentional decoder with the proposed selective-decoder, (\textbf{+0.4} pp and \textbf{+2.7} pp on regular and irregular text respectively).

Section (b) of \cref{tab:ablation} shows a monotonic increase in the accuracy using the SCATTER architecture with 4 BiLSTM layers by changing the number of intermediate supervisions (ranging from 1 to 3).
The relative increase in accuracy of section (b) is \textbf{+0.7} pp and \textbf{+2.9} pp on regular and irregular text respectively.
\subsection{Stable Training of a Deep BiLSTM Encoder}

As mentioned in the introduction, previous papers use only a 2-layer BiLSTM encoder.
The authors in~\cite{Zuo2019join} report a decrease in accuracy when increasing the number of layers in the BiLSTM encoder. 
We reproduce the experiment reported in~\cite{Zuo2019join} of training a baseline architecture with an increasing number of BiLSTM layers in the encoder (results are in the supplementary material).
We observe a similar phenomena as in~\cite{Zuo2019join}, i.e a reduction in accuracy when using more than two BiLSTM layers.
Contrary to this discovery, \cref{tab:ablation} shows that the overall trend of increasing the number of BiLSTM layers in \AlgoName, while increasing the number of intermediate supervisions improves accuracy.
Recognition accuracy improves monotonically up to 10 BiLSTM layers, both for regular and irregular text datasets.

As evident from \cref{tab:ablation} (c), when training with more than 10 BiLSTM layers in the encoder, accuracy results slightly (\textbf{-0.4} pp and \textbf{-0.5} pp on regular and irregular text respectively) decrease on both regular and irregular text (similar phenomena was observed on the validation set).
It is expected that increasing the network capacity leads to more challenging training procedures.
Other training approaches might need to be considered to successfully train to convergence a very deep encoder.
Such approaches may include incremental training, where we first train with a shallower network using a small number of blocks and incrementally stack more blocks during training.

\begin{table}
\normalsize
\begin{center}
\bgroup
\def\arraystretch{1.1}
\centering
  \begin{center}
    \caption{The table shows example for when repeated processing used in conjunction with intermediate supervision increasingly refines text predictions.}
    \label{tab:refinement_effect}
    \footnotesize
    \begin{tabular}{|c|P{0.5cm}P{0.5cm}P{0.5cm}P{0.6cm}|P{0.9cm}|P{0.9cm}|} 
      \hline
      \multirow{2}{*}{\textbf{Test Image}} & \multicolumn{4}{c|}{\textbf{Intermediate Decoder}} & \textbf{Final}& \textbf{Ground}\\
      & 1 & 2 & 3 & 4 & \textbf{Decoder} & \textbf{Truth} \\
      \hline
      \raisebox{-.4\height}{\includegraphics[width=1.4cm]{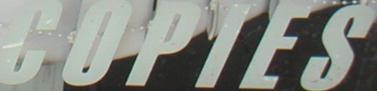}} &\textcolor{red}{g}\textcolor{green}{op}\textcolor{red}{t}\textcolor{green}{es} & \textcolor{green}{cop}\textcolor{red}{t}\textcolor{green}{es} & \textcolor{green}{copies} & \textcolor{green}{copies} & \textcolor{green}{copies} & copies\\
      &&&&&&\\
      \multirow{3}{*}{\includegraphics[width=1.4cm]{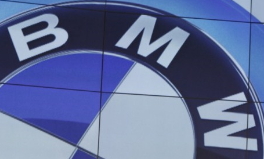}}  &  &  & & & &\\
      & \textcolor{green}{bm}\textcolor{red}{\_} & \textcolor{green}{b}\textcolor{red}{n\_} & \textcolor{green}{bm}\textcolor{red}{y} & \textcolor{green}{bm}\textcolor{red}{y} & \textcolor{green}{bmw} & bmw\\
      &&&&&&\\
      \raisebox{-.1\height}{\includegraphics[width=1.4cm]{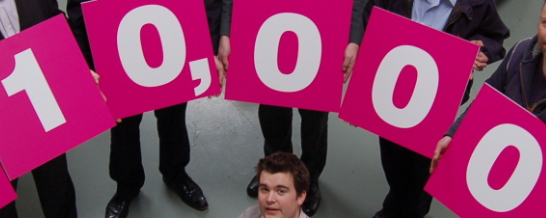}} & \textcolor{red}{o}\textcolor{green}{0}\textcolor{red}{\texttt{\_\_\_}}  & \textcolor{green}{100}\textcolor{red}{\texttt{\_\_}} & \textcolor{green}{100}\textcolor{red}{\texttt{\_\_}} & \textcolor{green}{1000}\textcolor{red}{\texttt{\_}} & \textcolor{green}{10000} & 10000\\
      \hline
    \end{tabular}
  \end{center}
\egroup
\end{center}
\end{table}

Examples of intermediate predictions are seen in \cref{tab:refinement_effect}, showcasing \AlgoName ability to increasingly refine text prediction. 

\subsection{Oracle Decoder Voting}
\label{subsec:oracle}
\begin{table}[t]
\normalsize
\begin{center}
\bgroup
\def\arraystretch{1.1}
\centering
  \begin{center}
  \captionsetup[table]{skip=10pt}
    \caption{Performance of each decoder and the oracle. The oracle performance is calculated given the ground truth.}
    \label{tab:decoder_limit}
    \footnotesize
    \begin{tabular}{|P{1.0cm}|P{0.70cm}P{0.50cm}P{0.50cm}P{0.50cm}|P{0.50cm}P{0.50cm}P{0.65cm}|} 
      \hline
      \textbf{Decoder} & \textbf{IIIT5K} & \textbf{SVT} & \textbf{IC03} & \textbf{IC13} & \textbf{IC15} & \textbf{SVTP} & \textbf{CUTE}\\
      \hline
      $CTC$ & 89.6 & 84.6 & 93.5 & 90.9 & 69.9 & 77.4 & 77.8 \\
      \hline
      $Attn_1$ & 93.5 & 90.3 & 95.9 & 93.9 & 82.9 & \dunderline{0.7pt}{86.4} &\textbf{ 87.5} \\
      \hline
      $Attn_2$ & 93.5 & 90.4 & \dunderline{0.7pt}{96.3} & 94.1 & 82.6 & 86.0 & 86.8 \\
      \hline
      $Attn_3$ & \textbf{93.7} & 91.0 & 95.7 & \dunderline{0.7pt}{94.3} & \dunderline{0.7pt}{82.7} & 85.1 & \dunderline{0.7pt}{87.2}\\
      \hline
      $Attn_4$ & \textbf{93.7} & 91.2 & \textbf{96.5} & \textbf{94.4} & \textbf{83.3} & 85.8 & 86.8\\
      \hline
      Final & \textbf{93.7} & \textbf{92.7} & \dunderline{0.7pt}{96.3} & 93.9 & 82.2 &  \textbf{86.9} & \textbf{87.5}\\
      \hline
      \hline
      Oracle & 96.3 & 95.1 & 97.1 & 95.5 & 87.1 & 89.9 & 90.3\\
      \hline
    \end{tabular}
  \end{center}
\egroup
\end{center}
\end{table}

In \cref{tab:decoder_limit} the test accuracy is shown for the intermediate decoders on a SCATTER architecture trained with 5 blocks.
The last row summarizes the potential results of an oracle, that for every test image chooses the correct prediction, if one exists in any of the decoders.
If an optimal strategy to select between the different decoders predictions exists, the results on \textit{all} datasets achieve a new state-of-the-art.
The possible improvement in accuracy achievable by such an oracle ranges between \textbf{+0.8} pp, and up to \textbf{+5} pp across the datasets.
A possible prediction selection strategy might be based on ensemble techniques, or a meta model that predicts which decoder to use for each specific image.

\section{Conclusions and Future Work}
\label{sec:conclusions}
In this work we propose a stacked block architecture named \AlgoName, which achieves SOTA recognition accuracy and enables stable, more robust training for STR networks that use deep BiLSTM encoders. 
This is achieved by adding intermediate supervisions along the network layers, and by relying on a novel selective decoder.

We also demonstrate that a repetitive processing architecture for text recognition, trained with intermediate selective decoders as supervision, increasingly refines text predictions.
In addition, other approaches of attention could also benefit from stacked attention decoders, as our proposed novelty is not limited to our formulation of attention.

We consider two promising directions for future work.
First, in \cref{fig:val_acc_heads} we show that training deeper networks and then pruning the last decoders (and layers) is preferable over training a shallower network.
This could lead to an increase in performance given a constraint on computational budget.
Finally, we see potential in developing an optimal selection strategy between the predictions of the different decoders for each image.

\clearpage

\appendix
\section*{Supplementary Materials}
\section{Regular Vs Irregular Text}
\begin{figure}
  \centering
  \includegraphics[width=\columnwidth]{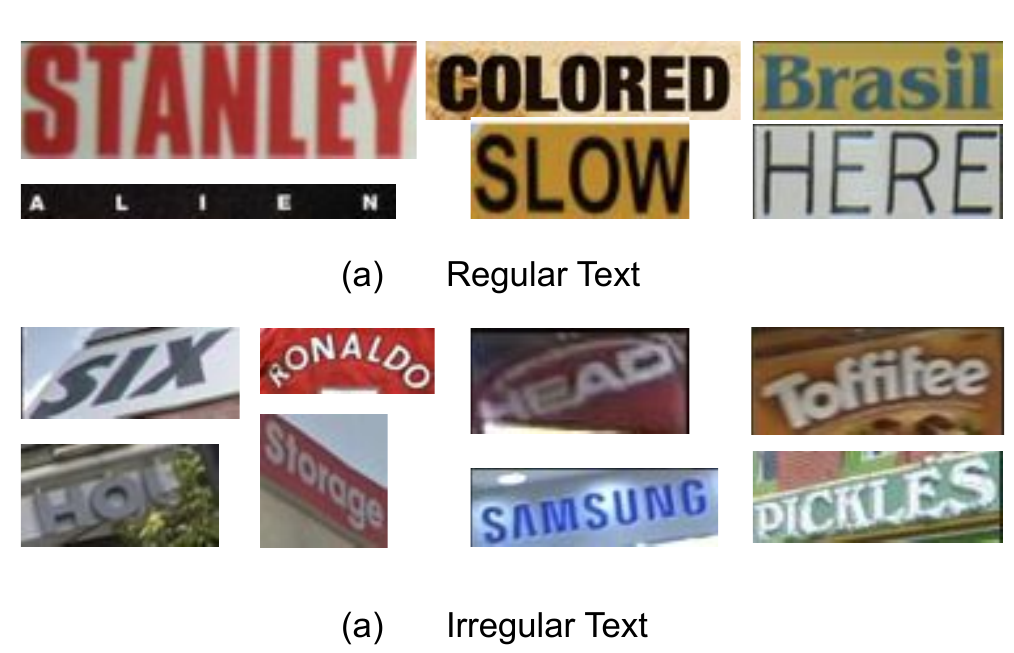}
  \caption{Examples of regular (IIIT5k, SVT, IC03, IC13) and irregular (IC15, SVTP, CUTE) real-world datasets.}
  \label{fig:sample_images}
\end{figure}

Recent works distinguish between two types of scene-text datasets:
\textit{Irregular text} where the text may be arbitrarily shaped (e.g. curved text), and \textit{regular text} where the sequence of characters is nearly horizontally aligned.
In \cref{fig:sample_images} we bring examples that demonstrate the main differences between these two types.

\section{Network Pruning - Compute Constraint}
\Cref{fig:val_acc_heads} in the main manuscript, shows accuracy levels for all intermediate decoders on several different stacking arrangements for training (e.g., using 1, 3, and 5 blocks).
\Cref{tab:prune} shows the exact results of \cref{fig:val_acc_heads}, with the additional results of all stacking arrangements for training, from a single block up to five blocks.
The results demonstrate that in general, it is favorable to train a deep network (with more blocks) and then prune, compared to training with a shallow architecture in the first place.
For example, if the target architecture inference time should include only 2 BiLSTM layers (similar to \cite{Baek2019clova}).
Training a 5-block SCATTER and pruning to a single block (11th row in \cref{tab:prune}) achieves \textbf{+0.4\%} pp and \textbf{+1.3\%} pp on regular and irregular text respectively, compared to training a single block (first row of the table) in the first place.

\section{Examples of Intermediate Predictions}
Following the discussion in \cref{subsec:oracle} of the paper, we provide additional examples of intermediate predictions for both regular and irregular text in \cref{tab:refinement_effect_supp}.
\Cref{tab:refinement_effect_supp} shows that in some cases (the first two rows for each text type) the earlier decoders fail to predict the word in the image, while the final decoder is correct.
In other cases (the last two rows for each text type) at least one of the intermediate decoders predicts the correct text, however, the final decoder fails to do so.
The described phenomena suggests that one could develop selection, voting or ensemble technique to improve results by choosing the correct prediction out of the available selective-decoders outputs.

\begin{table}[h]
\begin{center}
\small
\caption{Average test accuracy at intermediate decoding stages of the network, compared across different training network depths. * Regular Text and Irregular Text columns are weighted (by size) average results on the regular and irregular datasets respectively.}
\label{tab:prune}
\footnotesize
\bgroup
\def\arraystretch{1.1}
    \begin{tabular}{|P{1.0cm}|P{1.6cm}|P{1.6cm}|P{1.0cm}|P{1.0cm}|} 
      \hline
      \textbf{Training Blocks} & \textbf{N Blocks After Pruning} & \textbf{N LSTM Layers After Pruning} & \textbf{Regular Text*} & \textbf{Irregular Text*}\\
      \hline
      1 & 1 & 2 & 93.2 & 82.7 \\
      \hline
      \hline
      2 & 1 & 2 & 93.2 & 82.6 \\
      \hline
      2 & 2 & 4 & 93.6 & 83.0 \\
      \hline
      \hline
      3 & 1 & 2 & 93.8 & 83.2 \\
      \hline
      3 & 2 & 4 & 93.9 & 83.2 \\
      \hline
      3 & 3 & 6 & 93.7 & 83.4 \\
      \hline
      \hline
      4 & 1 & 2 & 93.4 & 83.5 \\
      \hline
      4 & 2 & 4 & 93.4 & 83.9 \\
      \hline
      4 & 3 & 6 & 93.6 & 83.4 \\
      \hline
      4 & 4 & 8 & 93.7 & 83.5 \\
      \hline
      \hline
      5 & 1 & 2 & 93.6 & 84.0 \\
      \hline
      5 & 2 & 4 & 93.7 & 83.7 \\
      \hline
      5 & 3 & 6 & 93.8 & 83.6 \\
      \hline
      5 & 4 & 8 & 94.0 & 84.1 \\
      \hline
      5 & 5 & 10 & 94.0 & 83.7 \\
      \hline
    \end{tabular}
\egroup
\end{center}
\end{table}

\begin{table*}[t]
\normalsize
\begin{center}
\bgroup
\def\arraystretch{1.1}
\centering
  \begin{center}
    \caption{Examples of intermediate decoders predictions on eight different images, from both regular and irregular text datasets. The presented results in the table, suggests that a selection, voting or ensemble technique could be use to improve results}
    \label{tab:refinement_effect_supp}
    \footnotesize
    \begin{tabular}{|c|c|P{1.5cm}P{1.5cm}P{1.5cm}P{1.5cm}|P{1.7cm}|P{1.7cm}|} 
      \hline
      \multirow{2}{*}{\textbf{Text Type}} & \multirow{2}{*}{\textbf{Test Image}} & \multicolumn{4}{c|}{\textbf{Intermediate Decoder}} & \textbf{Final}& \textbf{Ground}\\
      & & 1 & 2 & 3 & 4 & \textbf{Decoder} & \textbf{Truth} \\
      \hline
      \multirow{10}{*}{Reguler} & \multirow{3}{*}{\raisebox{-.3\height}{\includegraphics[width=1.4cm]{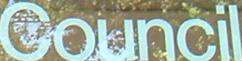}}}  &  &  & & & &\\
      & & \textcolor{red}{o}\textcolor{green}{ouncil} & \textcolor{red}{o}\textcolor{green}{ouncil} & \textcolor{green}{council} & \textcolor{green}{council} & \textcolor{green}{council} & council\\
      &&&&&&&\\
      &\multirow{3}{*}{\raisebox{-.3\height}{\includegraphics[width=1.4cm]{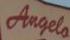}}}  &  &  & & & &\\
      && \textcolor{green}{angel}\textcolor{red}{s} & \textcolor{green}{angel}\textcolor{red}{s} & \textcolor{green}{angel}\textcolor{red}{s} & \textcolor{green}{angelo} & \textcolor{green}{angelo} & angelo\\
      &&&&&&&\\
      &\raisebox{-.1\height}{\includegraphics[width=1.4cm]{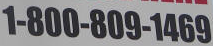}} & \textcolor{green}{18008091469}  & \textcolor{green}{18008091469} & \textcolor{green}{18008091469} & \textcolor{green}{18008091469} & \textcolor{green}{1800809146}\textcolor{red}{6}\textcolor{green}{9} & 18008091469\\
      &\multirow{3}{*}{\includegraphics[width=1.4cm]{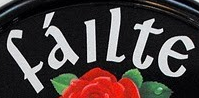}}  &  &  & & & &\\
      && \textcolor{red}{l}\textcolor{green}{ailte} & \textcolor{green}{failte} & \textcolor{green}{failte} & \textcolor{red}{l}\textcolor{green}{ailte} & \textcolor{red}{l}\textcolor{green}{ailte} & failte\\
      &&&&&&&\\
      \hline
      \multirow{14}{*}{Irreguler} & \multirow{3}{*}{\raisebox{-.3\height}{\includegraphics[width=1.4cm]{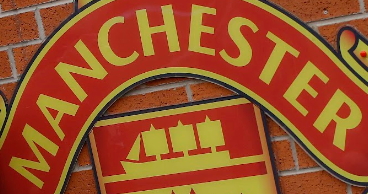}}}  &  &  & & & &\\
      & & \textcolor{red}{an}\textcolor{green}{nchester} & \textcolor{red}{wi}\textcolor{green}{nchester} & \textcolor{red}{w}\textcolor{green}{anchester} & \textcolor{red}{b}\textcolor{green}{anchester} & \textcolor{green}{manchester} & manchester\\
      &&&&&&&\\
      &\multirow{3}{*}{\includegraphics[width=1.4cm]{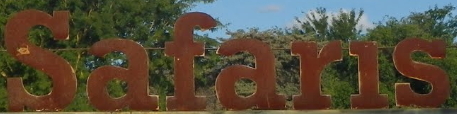}}  &  &  & & & &\\
      && \textcolor{green}{shant}\textcolor{red}{h}\textcolor{green}{a}\textcolor{red}{nt} & \textcolor{green}{safaris} & \textcolor{green}{safaris} & \textcolor{green}{safari}\textcolor{red}{c} & \textcolor{green}{safaris} & safaris\\
      &&&&&&&\\
      &\raisebox{-.1\height}{\includegraphics[width=1.4cm]{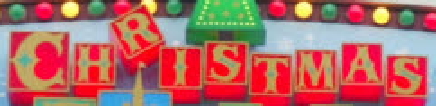}} & \textcolor{green}{ch}\textcolor{red}{\_\_}\textcolor{green}{stmas}  & \textcolor{green}{christmas} & \textcolor{green}{christ}\textcolor{red}{in}\textcolor{green}{as} & \textcolor{green}{christ}\textcolor{red}{w}\textcolor{green}{as} & \textcolor{green}{christ}\textcolor{red}{w}\textcolor{green}{as} & christmas\\
      &\multirow{7}{*}{\includegraphics[width=1.4cm]{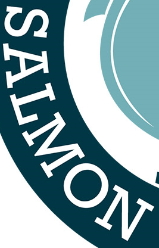}}  &  &  & & & &\\
      &&&&&&&\\
      &&&&&&&\\
      && \textcolor{red}{b}\textcolor{green}{almon} & \textcolor{green}{salmon} & \textcolor{green}{salmon} & \textcolor{red}{b}\textcolor{green}{almon} & \textcolor{red}{b}\textcolor{green}{almon} & salmon\\
      &&&&&&&\\
      &&&&&&&\\
      &&&&&&&\\
      \hline
    \end{tabular}
  \end{center}
\egroup
\end{center}
\end{table*}

\section{Stable Training of a Deep BiLSTM Encoder}

\begin{table}[ht]
\begin{center}
\footnotesize
\caption{The effect of the number of BiLSTM layers used on recognition accuracy. Only by using SCATTER we are able to add BiLSTM layer to improve results. Regular Text and Irregular Text columns are weighted (by size) average results on the regular and irregular datasets.
We refer to our re-trained model using the code of Baek et al. 2019 as Baseline. }
\label{tab:lstm}
\footnotesize
\bgroup
\def\arraystretch{1.1}
    \begin{tabular}{|P{1.3cm}|P{1.7cm}|P{1.6cm}|P{1.8cm}|} 
      \hline
      \textbf{\# Blocks} & \textbf{LSTM Layers} & \textbf{Regular Text} & \textbf{Irregular Text}\\
      \hline
      Baseline & 1 & 92.5 & 79.0 \\
      \hline
      Baseline & 2 & 92.7 & 79.1 \\
      \hline
      Baseline & 3 & 92.6 & 78.7 \\
      \hline
      Baseline & 4 & 92.4 & 78.6 \\
      \hline
      \hline
      1 & 2 & 93.2 & 82.7 \\
      \hline
      2 & 4 & 93.6 & 83.0 \\
      \hline
      3 & 6 & 93.7 & 83.4 \\
      \hline
      4 & 8 & 93.7 & 83.5 \\
      \hline
      5 & 10 & 94.0 & 83.7 \\
      \hline
    \end{tabular}
\egroup
\end{center}
\end{table}
Previous papers used only a 2-layer BiLSTM encoder.
In~\cite{Zuo2019join} the authors report a decrease in accuracy while increasing the number of layers in the BiLSTM encoder.
In \cref{tab:lstm} we show results of a reproduction of the experiment reported in~\cite{Zuo2019join}, training a baseline architecture with an increasing number of BiLSTM layers in the encoder.
We observe a similar phenomena to~\cite{Zuo2019join} -- a reduction in accuracy when using more than two BiLSTM layers in the baseline architecture.
However, the bottom rows of the table demonstrate that SCATTER allows stacking of more BiLSTM layers, which ultimately leads to an increase in final performance.

\end{document}